UDC 004.056.5

# RESEARCH ON A HYBRID LSTM-CNN-ATTENTION MODEL FOR TEXT-BASED WEB CONTENT CLASSIFICATION

**Kuz M. V.** – Dr. Sc., Professor of the Department of Information Technology Vasyl Stefanyk Precarpathian National University, Ukraine.

**Lazarovych I. M.** – PhD, Associate Professor of the Department of Information Technology Vasyl Stefanyk Precarpathian National University, Ukraine.

**Kozlenko M. I.** – PhD, Associate Professor of the Department of Information Technology Vasyl Stefanyk Precarpathian National University, Ukraine.

**Pikuliak M. V.** – PhD, Associate Professor of the Department of Information Technology Vasyl Stefanyk Precarpathian National University, Ukraine.

**Kvasniuk A. D.** – Graduate student, Vasyl Stefanyk Precarpathian National University, Ukraine.

## ABSTRACT

**Context.** Text-based web content classification plays a pivotal role in various natural language processing (NLP) tasks, including fake news detection, spam filtering, content categorization, and automated moderation. As the scale and complexity of textual data on the web continue to grow, traditional classification approaches – especially those relying on manual feature engineering or shallow learning techniques – struggle to capture the nuanced semantic relationships and structural variability of modern web content. These limitations result in reduced adaptability and poor generalization performance on real-world data. Therefore, there is a clear need for advanced models that can simultaneously learn local linguistic patterns and understand the broader contextual meaning of web text.

**Method.** This study presents a hybrid deep learning architecture that integrates Long Short-Term Memory (LSTM) networks, Convolutional Neural Networks (CNN), and an Attention mechanism to enhance the classification of web content based on text. Pre-trained GloVe embeddings are used to represent words as dense vectors that preserve semantic similarity. The CNN layer extracts local $n$-gram patterns and lexical features, while the LSTM layer models long-range dependencies and sequential structure. The integrated Attention mechanism enables the model to focus selectively on the most informative parts of the input sequence. The model was evaluated using the dataset, which consists of over 10,000 HTML-based web pages annotated as legitimate or fake. A 5-fold cross-validation setup was used to assess the robustness and generalizability of the proposed solution.

**Results.** Experimental results show that the hybrid LSTM-CNN-Attention model achieved outstanding performance, with an accuracy of 0.98, precision of 0.94, recall of 0.92, and F1-score of 0.93. These results surpass the performance of baseline models based solely on CNNs, LSTMs, or transformer-based classifiers such as BERT. The combination of neural network components enabled the model to effectively capture both fine-grained text structures and broader semantic context. Furthermore, the use of GloVe embeddings provided an efficient and effective representation of textual data, making the model suitable for integration into systems with real-time or near-real-time requirements.

**Conclusions.** The proposed hybrid architecture demonstrates high effectiveness in text-based web content classification, particularly in tasks requiring both syntactic feature extraction and semantic interpretation. By combining convolutional, recurrent, and attention-based mechanisms, the model addresses the limitations of individual architectures and achieves improved generalization. These findings support the broader use of hybrid deep learning approaches in NLP applications, especially where complex, unstructured textual data must be processed and classified with high reliability.

## ABBREVIATIONS

CNN is a Convolutional Neural Network;

LSTM is a Long Short-Term Memory;

GloVe is a set of Global Vectors for Word Representation;

AUC is an Area Under the Curve.

## NOMENCLATURE

$D$ is a set of documents (e.g., HTML content of web pages);

$f$ is a classification function mapping documents or features to predicted classes;

$R^m$ is a feature space after text preprocessing;

$\theta$ is a set of trainable model parameters;

$L(\theta)$ is a loss function measuring prediction error;

$\theta^*$ is an optimal parameter set minimizing the loss function;

*True Positive (TP)* is a number of web pages that the model correctly classified as belonging to the target category (e.g., fake news or spam);

*False Positive (FP)* is a number of web pages that the model incorrectly classified as belonging to the target category when they do not;

*False Negative (FN)* is a number of web pages that the model incorrectly classified as not belonging to the target category when they actually do;

*True Negative (TN)* is a number of web pages that the model correctly classified as not belonging to the target category;

*TPR* is a True Positive Rate;

*FPR* is a False Positive Rate, which are plotted on the ROC curve;

$x^{(s)}$ is a $s$-th input text sample;

$w^{(t)}$ is a $t$-th word in the sequence;

$E$ is a pre-trained embedding matrix;

$V$ is a vocabulary size;

$d$ is a dimensionality of the word embeddings;

$e_t$ is an embedding vector of the word $w_t$;

$X^{(s)}$ is a matrix of embeddings for sample $s$;

$f$ (in CNN layer) is a convolutional filter;

$k$ is a kernel width of the filter;

$X_{i:i+k-1}$ is a sliding window input over the embeddings;

$b$ is a bias term in CNN;

$\langle \cdot, \cdot \rangle$ is an element-wise product followed by summation;

$c_i$ is a feature after convolution and activation;

$C$ is a feature map;

$\hat{c}$ is a result of max-pooling;

$x_t$ is an input vector at time step $t$ in the LSTM;

$h_t$ is a cell state at time step $t$;

$\sigma(\cdot)$ is a sigmoid activation function;

$\circ$ is an element-wise (Hadamard) product;

$i_t, f_t, o_t$ are input, forget, and output gates in the LSTM;

$\tilde{c}_t$ is a candidate cell state;

$W_i, W_f, W_o, W_c$ are weight matrices for the input;

$U_i, U_f, U_o, U_c$ are recurrent weight matrices;

$b_i, b_f, b_o, b_c$ are bias vectors for each gate;

$H$ is a matrix of LSTM hidden states;

$u$ is a context vector in the attention mechanism;

$\alpha_t$ is an attention weight for time step $t$;

$c$ (in attention) is a context vector, a weighted sum of hidden states;

$\hat{s}$ is a predicted probability distribution over classes;

$W_c$ is a weight matrix in the classification layer;

$b_c$ is a bias term in the classification layer.

## INTRODUCTION

The explosive growth of web-based textual data has led to increasing demand for intelligent systems capable of automatically classifying web content. Tasks such as fake news detection, spam filtering, opinion mining, and automated content moderation rely heavily on accurate and scalable classification methods. A major challenge in this domain is the diversity and complexity of natural language used across different types of web content, which often includes both structured and unstructured text embedded in HTML pages.

Traditional classification systems used to detect such deceptive content often rely on manually designed features, blacklists, or rule-based heuristics [1]. While these approaches can be effective in specific, known scenarios, they typically struggle to generalize to new or sophisticated cases, especially as the structure and language of web content continue to evolve. Their limited adaptability and dependence on static features reduce their effectiveness in large-scale or dynamic environments.

In contrast, modern deep learning approaches [2] – particularly those incorporating Convolutional Neural Networks (CNN), Long Short-Term Memory (LSTM) networks, and Attention mechanisms-offer a powerful alternative. These models can learn meaningful patterns directly from raw textual input, capturing both local structures and long-range dependencies. They are well suited for web content classification tasks that require the extraction of complex semantic and syntactic features from natural language.

**The object of study** is the process of classifying text-based web content using a hybrid deep learning model that integrates CNN, LSTM, and Attention components. The model is trained on labeled HTML-based web pages that include both legitimate and deceptive content, with the goal of automatically assigning them to the correct class.

**The subject of study** is the development of a hybrid neural architecture that combines the strengths of CNNs, LSTMs, and Attention mechanisms for improved classification of web-based textual data.

**The purpose of the work** is to enhance the effectiveness and generalization capabilities of web content classification models by leveraging advanced deep learning techniques. The aim is to improve both feature extraction and sequence modeling, ultimately leading to a more robust and accurate system for identifying deceptive or misleading textual web content.

## 1 PROBLEM STATEMENT

The classification of web content based on textual information is a fundamental task in natural language processing. It plays a critical role in applications such as topic categorization [3], content moderation [4], spam detection, and personalized content delivery. Given the large volume and unstructured nature of online data, there is a growing need for intelligent models that can automatically analyze and classify web content with high accuracy.

The web content classification task can be formulated as a supervised multi-class or binary classification problem over a set of documents. Let:

– $D = \{d_1, d_1, ..., d_n\}$ as a set of documents (HTML content of web pages);

– where each $d_i \in D$ has an associated label $y_i \in \{0,1\}$, with 0 representing a standard (legitimate) page and 1 representing a target class of interest.

The objective is to find a classification function (1):

$$f : D \rightarrow \{0,1\}. \qquad (1)$$

Since HTML documents are textual in nature, they are typically preprocessed and transformed into numerical representations before being used in classification models.

OPEN ACCESS

Therefore, the problem can be equivalently defined in the feature space as (2):

$$f : R^m \rightarrow \{0,1\}. \qquad (2)$$

The goal is to optimize the classification function's parameters $\theta$ in such a way that a chosen loss function $L(\theta)$ is minimized (3):

$$\theta^* = \arg\min_{\theta} L(\theta). \qquad (3)$$

## 2 REVIEW OF THE LITERATURE

Initial methods for web content classification were rule-based or relied on classic machine learning models, such as Naïve Bayes [5], Support Vector Machines (SVM) [5], and Decision Trees [5]. These models worked by extracting features from web pages, such as URL characteristics and HTML content. Despite their utility, these models had several limitations. They lacked contextual awareness of the text, and their performance degraded when handling large datasets or sophisticated content structures.

For example, Naïve Bayes and SVMs were constrained by the simplicity of the features they utilized, often failing to capture the complex structure of web pages. Moreover, these models required extensive manual tuning and were prone to misclassifying web pages that did not exhibit obvious, rule-based features.

With the advent of natural language processing (NLP) techniques, newer methods based on word embeddings, such as TF-IDF [6], Word2Vec [6], and GloVe, began to emerge. These methods improved traditional feature extraction by capturing the semantic meaning of words and their contextual relationships [7]. However, while TF-IDF represented a significant improvement, it lacked the ability to account for word order, which was crucial for understanding the structure of the content. Word2Vec enhanced this by embedding words in continuous vector spaces that preserved their meanings in context. Despite this, Word2Vec still struggled with understanding the global context of the text, making it less effective for long-form web pages or complex content.

GloVe addressed some of these challenges by factorizing the word co-occurrence matrix to capture global word relationships. However, as noted in the study, GloVe's use of pre-trained word vectors allowed for faster training and reduced computational complexity compared to models like BERT, which required significant resources for fine-tuning [8].

Recent advances in deep learning have significantly impacted web content classification. Convolutional Neural Networks (CNNs) [9, 10] and Long Short-Term Memory (LSTM) networks [11] have been successfully applied to text classification tasks due to their ability to model both local and sequential dependencies, respectively. CNNs are particularly effective for detecting local patterns, such as identifying suspicious phrases or structural features in web pages. However, CNNs struggle with capturing long-term dependencies in textual data.

In contrast, LSTMs excel at modeling sequential data and are better at preserving the long-term context in web content. While LSTMs are more effective at handling longer sequences, they can still encounter difficulties in modeling very long-term dependencies and require significant training time.

The introduction of transformer-based models, such as BERT [12], marked a new era in text classification. BERT, with its attention mechanism, is capable of understanding the context of words within a sentence by considering the relationships between all words in the text. This makes it highly effective for understanding complex, context-dependent content. However, transformer models are computationally expensive and require fine-tuning on domain-specific data to achieve optimal results.

Although BERT shows superior performance in tasks like web content classification, its high computational cost has prompted researchers to explore lighter alternatives. For instance, the hybrid CNN-LSTM-Attention model proposed in this study offers a good balance between performance and computational efficiency. This model combines the strengths of CNNs and LSTMs, with the addition of an Attention mechanism that highlights the most critical parts of the text, improving both the interpretability and accuracy of the model.

## 3 MATERIALS AND METHODS

For web content classification in this study, a dataset containing web pages in HTML code format, labeled as either "phishing" or "legitimate", was used. This dataset [13] is publicly available and was originally presented as an SQL archive. To facilitate its usage, the data was extracted, opened in the HeidiSQL environment, and exported to the CSV format, which is more convenient for processing in Python.

The dataset consists of three tables, but only the websites table was used for this research. Two key columns were selected from it:

– htmlContent, which contains the HTML code of the web pages;

– isPhish, which indicates whether a page contains deceptive web content or legitimate content.

After preprocessing, which includes text cleaning, lemmatization, bigram generation, and tokenization, the Phishload dataset contains 10.373 records, of which 9.198 are deceptive web pages and 1.176 are legitimate ones. This distribution highlights a significant class imbalance, which may affect model training, as the majority of samples belong to one class.

Fig. 1 illustrates the class distribution in this dataset, showing the dominance of deceptive web pages over legitimate ones, which necessitates the use of specialized balancing techniques.

Fig. 2 illustrates the distribution of HTML code length in the Phishload dataset, allowing for an assessment of the differences between deceptive and legitimate web pages.

Text preprocessing is a crucial step in machine learning and natural language processing tasks [14]. It enables the transformation of textual data into a more structured and analyzable format, significantly improving the efficiency of classification algorithms. This is particularly relevant for harmful content detection, as fraudulent messages often contain word variations, grammatical distortions, and excessive information, complicating analysis.

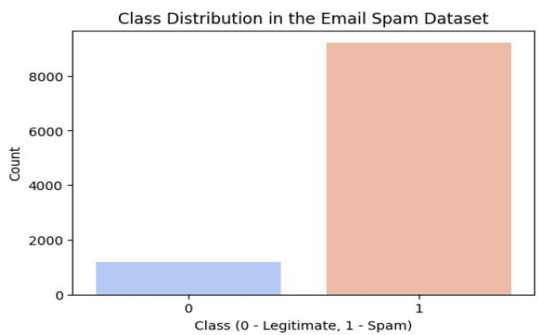

Figure 1 – The class distribution in the HTML dataset

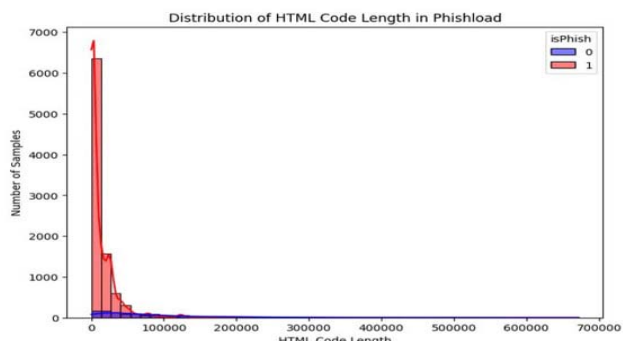

Figure 2 – The distribution of HTML code length in the dataset

This study examines key text preprocessing methods used for analyzing and classifying web content. The main methods include lemmatization, tokenization, stop-word removal, and bigram formation. These techniques enhance the quality of textual data, positively impacting the accuracy of classification models.

Tokenization splits text into smaller units (tokens). Deceptive web pages may use obfuscation (e.g., 'p@ssw0rd') to evade detection. NLTK [15] ensures precise segmentation. Lemmatization reduces words to their base form (e.g., 'running' → 'run') to minimize redundancy. spaCy was used for effective context-aware processing.

Stop-word removal eliminates common words ('the', 'is', etc.) to retain only relevant terms. Harmful web content often includes manipulative phrases ('urgent', 'click here'), which were filtered using NLTK.

Bigrams capture meaningful word pairs ("credit_card") to enhance context understanding. The generate_ngrams function created bigrams for improved text analysis.

Applying tokenization, lemmatization, stop-word removal, and bigram formation significantly improves the quality of textual data before passing it into a machine learning model. This, in turn, enhances the efficiency of classification algorithms, allowing for more accurate detection of harmful content.

Text vectorization [16] is a crucial step in NLP tasks, as textual data must be transformed into a format that machine learning algorithms can process.

Fig. 3 illustrates the sequence of key text preprocessing steps before feeding data into a machine learning model.

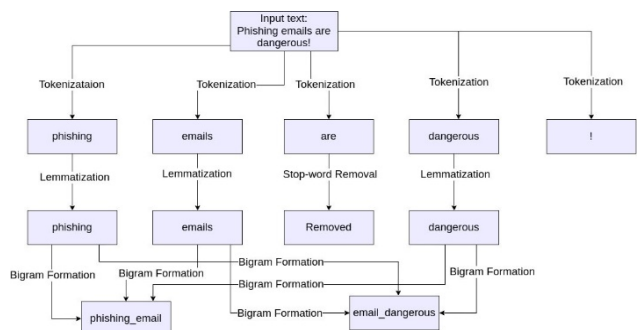

Figure 3 – General process of text preprocessing for web content classification

In this study, a comparative analysis of key vectorization approaches was conducted using several evaluation metrics to assess their effectiveness in the context of web content classification. Specifically, the models were compared based on the following criteria:

Precision and recall. These metrics were used to evaluate the effectiveness of each vectorization method in correctly identifying harmful content. Precision and recall are defined as (4, 5):

$$Precision = \frac{True\ Positives}{True\ Positives + False\ Positives}. \quad (4)$$

$$Recall = \frac{True\ Positives}{True\ Positives + False\ Negatives}. \quad (5)$$

F1 Score. The F1 score was calculated as (6):

$$F1 = 2 \times \frac{Precision \times Recall}{Precision + Recall}. \quad (6)$$

Accuracy. The overall classification accuracy was computed as (7):

$$Accuracy = \frac{True\ Positives + True\ Negatives}{Total\ Samples}. \quad (7)$$

In addition to the above metrics, AUC was specifically utilized to further evaluate and compare the overall classification performance of the models, providing insight into their ability to distinguish between relevant and irrelevant web content across different classification thresholds.

The AUC score reflects the model's ability to separate positive (target category) and negative (non-target category) classes over all possible classification thresholds. It is computed as the area under the ROC curve (Receiver Operating Characteristic curve), which plots the true positive rate against the false positive rate, as (8):

$$AUC = \int_0^1 TRP(t) d(FPR(t)).$$  (8)

The following methods were employed for text vectorization:

– TF-IDF – a classical method that evaluates term importance in a document.

– Word2Vec – a neural network-based method that generates vector representations based on context.

– GloVe – a statistical method that captures word co-occurrences within a corpus.

– BERT – a transformer-based approach for contextual text understanding.

TF-IDF evaluates term frequency and is known for its simplicity and interpretability; however, it does not consider the context of words. Word2Vec trains a neural network based on word context, allowing it to preserve semantic relationships, but it requires a large corpus for effective training. GloVe factorizes a word co-occurrence matrix [17], making it well-suited for capturing word analogies, yet it demands significant computational resources. BERT, which utilizes transformers, achieves high accuracy in complex tasks but comes with a high computational cost.

A hybrid approach is proposed that combines CNN, LSTM, and an Attention mechanism, leveraging the strengths of each architecture. CNN effectively extracts local patterns in the text, allowing it to identify potentially suspicious phrases and structures. LSTM retains long-term context, which is crucial for analyzing sequences and understanding the overall meaning of a message. The Attention mechanism [18–19] further enhances the model by focusing on the most important words in the text, strengthening relevant features for classification.

The architecture of the hybrid model consists of several key stages. First, the input text is processed through a vectorization layer using GloVe. Next, CNN detects local patterns and key textual features, after which LSTM processes the obtained representations to preserve sequential context. The Attention mechanism identifies the most significant words that are critical for classification, and the final output is passed to a fully connected layer for decision-making.

The combination of local and global text analysis results in a more accurate model for web content classification. Integrating CNN and LSTM with the Attention mechanism enhances performance compared to using each method individually. This approach improves the model's ability to capture subtle linguistic patterns and semantic context, making it more adaptable and effective for handling diverse and complex textual data in real-world applications.

To formalize the operation of the proposed CNN-LSTM-Attention architecture, a mathematical description of each component is presented below.

Let $x^{(s)} = [w_1, w_2, ..., w_T]$ denote the $s$-th text sample, consisting of $T$ words.

To convert words into dense numerical representations, a pre-trained word embedding matrix $E \in R^{V*d}$ is used [20]. The resulting embedding sequence is (9):

$$X^{(s)} = [e_1, e_2, ..., e_T] \in R^{T \cdot d}, \; e_t = E(w_t).$$  (9)

The convolutional layer is used to capture local patterns in the text [21]. Let $f \in R^{k \cdot d}$ denote a convolutional filter of width $k$. The convolution operation over a window $X_{i:i+k-1}$ is defined as (10):

$$c_i = ReLU(\langle f, X_{i:i+k-1} \rangle) + b.$$  (10)

A feature map is formed as (11):

$$C = [c_1, c_2, ..., c_{T-k+1}].$$  (11)

To reduce dimensionality, a max-pooling operation is applied (12):

$$\hat{c} = \max(C).$$  (12)

The Long Short-Term Memory (LSTM) layer models long-range dependencies in the sequence [22]. The input is represented as (13):

$$X = [x_1, ..., x_T].$$  (13)

For each time step $t$, the LSTM computes (14):

$$
\begin{aligned}
i_t &= \sigma(W_i x_t + U_i h_{t-1} + b_i), \\
f_t &= \sigma(W_f x_t + U_f h_{t-1} + b_f), \\
o_t &= \sigma(W_o x_t + U_o h_{t-1} + b_o), \\
\tilde{c}_t &= \tanh(W_c x_t + U_o h_{t-1} + b_o), \\
c_t &= f_t \circ c_{t-1} + i_t \circ \tilde{c}_t, \\
h_t &= o_t \circ \tanh(c_t).
\end{aligned}
$$  (14)

To focus on the most informative parts of the sequence, a soft attention mechanism is applied. Let $H = [h_1, h_2, ..., h_T]$ be the sequence of LSTM hidden states [23]. The attention weight $\alpha_t$ for each time step is computed as (15):

$$\alpha_t = \frac{\exp(h_t^T u)}{\sum_{j=1}^{T} \exp(h_t^T u)}. \tag{15}$$

The context vector $c$, which summarizes the weighted hidden states, is calculated as (16):

$$c = \sum_{t=1}^{T} \alpha_t \times h_t. \tag{16}$$

The resulting context vector $c$, is passed through a fully connected layer with a softmax activation to obtain class probabilities (17):

$$\hat{s} = soft\max(W_c c + b_c), \ \hat{s} \in R^2. \tag{17}$$

The model is trained by minimizing the cross-entropy loss function [24] (18):

$$L = -\sum_{s=1}^{N} y^{(s)} \times \log(\hat{s}^{(s)}). \tag{18}$$

Fig. 4 illustrates the architecture of the CNN, LSTM, and Attention mechanism.

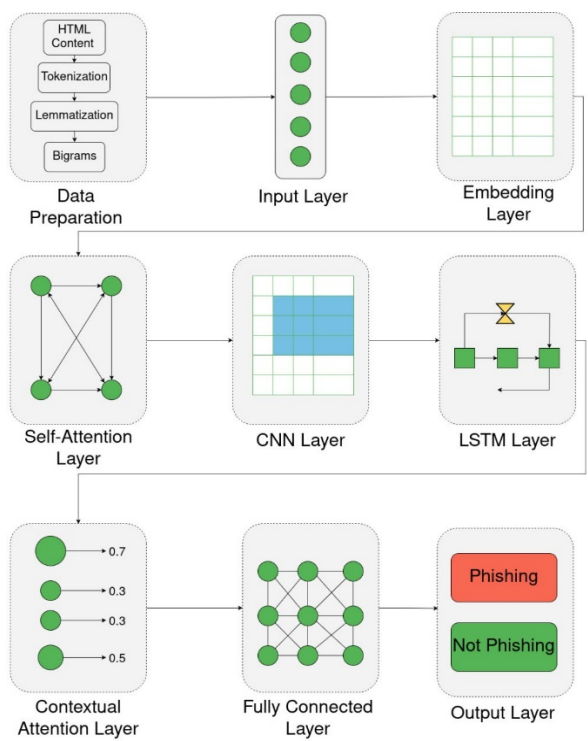

Figure 4 – Architecture of a Text-Based Web Content Classification Model Using CNN, LSTM, and Attention

At the initial stage, the model receives input data in the form of an HTML document, which has been preprocessed into a sequence of 200 tokens. This ensures a uniform input length, which is a prerequisite for further neural processing. Tokenization typically retains only the most frequent elements (e.g., top-5000).

Each input token is mapped into a 100-dimensional vector using pre-trained GloVe embeddings. This enables the model to preserve semantic relationships between words before training even begins. The parameter trainable=False ensures that the embeddings remain unchanged during training, promoting stability and generalization.

Once the tokens are embedded, convolutional operations are applied to extract local patterns in the HTML sequence. A 1D convolution with 128 filters of size 5 captures characteristic fragments of code (e.g., tag patterns or suspicious sequences of attributes). MaxPooling with a pool size of 5 reduces the dimensionality while retaining the most salient features.

To model long-range dependencies in HTML documents, the architecture employs a multi-head self-attention mechanism. This allows the model to attend to key structural elements, regardless of their position in the document. For example, a `<form>` tag at the beginning may be closely related to a `<submit>` button at the end. Functionally, this component is described as `Multi-HeadAttention(num_heads=4, key_dim=64)`.

To retain sequential context, the model incorporates a Long Short-Term Memory (LSTM) network. LSTM captures both positional and structural dependencies in HTML, which is especially critical for processing nested tags. The `return_sequences=True` setting ensures that the entire sequence is preserved for the next attention mechanism.

On top of the LSTM outputs, a trainable contextual attention mechanism is applied. This layer computes the context vector by focusing on the most relevant parts of the sequence, effectively summarizing the input into a weighted representation for final classification.

To mitigate overfitting, the model applies a Dropout layer with a dropout rate of 30%, randomly deactivating neurons during training. This improves the robustness and generalization ability of the model.

The final layer is a dense [25] classifier with a softmax activation function, which produces a probability distribution over the two output classes: relevant (1) or irrelevant (0) web content.

## 4 EXPERIMENTS

For deceptive content classification, an experiment was conducted to assess model performance. K-Fold cross-validation [26] with K = 5 was applied, splitting the dataset into 80% for training and 20% for testing. The hyperparameters set are as follows: embedding dimension is 100, LSTM units are 128, optimizer is adam, learning rate is 0.001, batch size is 32, and the number of epochs is 20. This approach ensures a more objective evaluation, reduces dependency on specific dataset partitions, and improves the reliability of the results.

Among the four text vectorization methods (BERT, GloVe, Word2Vec, TF-IDF), the GloVe method was selected for the following reasons:

Stable High Accuracy. The model with GloVe quickly achieves consistently high accuracy on both training and validation data (Fig. 5). This indicates that GloVe effectively extracts key semantic features of the text.

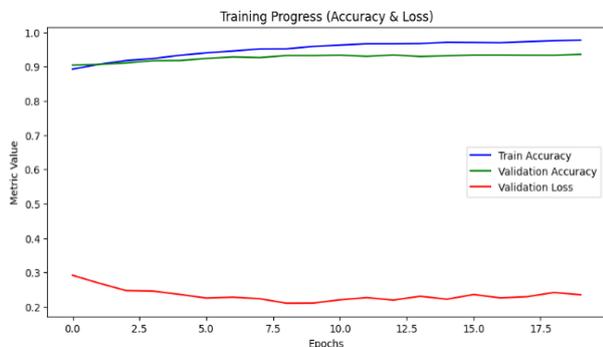

Figure 5 – The training process: changes in accuracy and loss

Fig. 6–7 present the results after performing K-Fold cross-validation for CNN, LSTM, and the Attention mechanism.

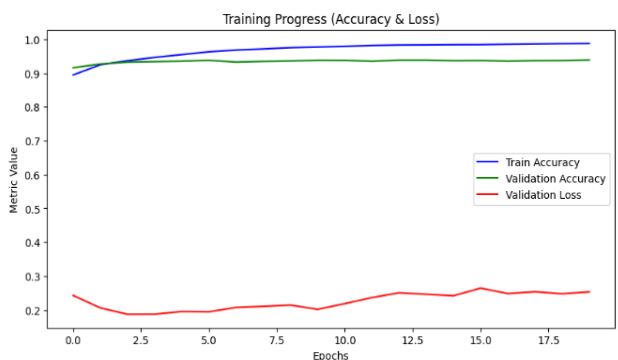

Figure 6 – Training process after performing K-Fold cross-validation for CNN, LSTM, and the Attention mechanism: changes in accuracy and loss

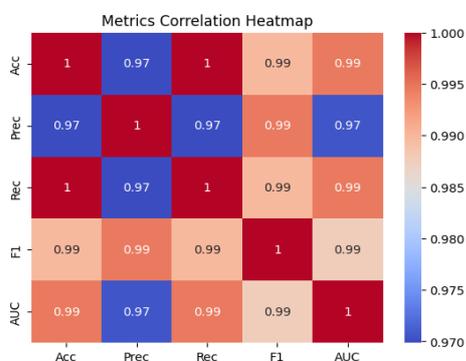

Figure 7 – Correlation analysis between Accuracy, Precision, Recall, F1, and AUC after performing K-Fold cross-validation for CNN, LSTM, and the Attention mechanism

The presented results demonstrate that the model maintains consistently high accuracy on both training and validation data, effectively distinguishing between deceptive and legitimate messages. This is also confirmed by the balance of the Accuracy, Precision, Recall, F1-score, and AUC metrics, indicating the model's stability without bias toward any particular class.

## 5 RESULTS

The performance metrics presented in tables were obtained using the dataset [13], which contains web pages in HTML code format labeled as either "phishing" or "legitimate".

From Table 1, we can observe the performance of different methods in harmful content classification based on accuracy, precision, recall, and F1-score.TF-IDF shows the lowest performance across all metrics, with an accuracy of 0.88 and an F1-score of 0.61, indicating its limitations in capturing contextual relationships.Word2Vec and GloVe outperform TF-IDF, achieving 0.93 and 0.94 accuracy, respectively. Both methods maintain high precision (0.87 and 0.88) and recall (0.87 and 0.86), resulting in an F1-score of 0.87 for both. This suggests that these word embedding techniques effectively capture semantic relationships. BERT achieves an accuracy of 0.92, comparable to Word2Vec and GloVe. However, while its precision (0.88) is on par with GloVe, it has a lower recall (0.79), leading to a slightly lower F1-score (0.81). This could indicate that BERT's performance is affected by the dataset size or training constraints. Overall, GloVe provides the highest accuracy (0.94) and maintains a strong balance between precision and recall, making it the best-performing method in this comparison.

From Table 2, the results indicate that the model achieves high accuracy and strong generalization ability. The addition of the attention mechanism improves the extraction of key textual features, enhancing classification quality. Fig. 7 illustrates the balance between accuracy, precision, recall, F1-score, and AUC, which is essential for deceptive content detection. The combination of CNN and LSTM effectively captures both local and global dependencies in text, reducing the gap between training and validation accuracy (Fig. 6). While the model requires more resources than the baseline LSTM, it is significantly faster than BERT, making it an optimal choice for harmful content classification.

Table 1 – Comparison of text preprocessing methods for risky content analysis

| Method | Accuracy | Precision | Recall | F1-score |
|--------|----------|-----------|--------|----------|
| TF-IDF | 0.88 | 0.59 | 0.65 | 0.61 |
| Word2Vec | 0.93 | 0.87 | 0.87 | 0.87 |
| GloVe | 0.94 | 0.88 | 0.86 | 0.87 |
| BERT | 0.92 | 0.88 | 0.79 | 0.81 |

Table 2 – Metrics of the trained models

| Method | Accuracy | Precision | Recall | F1-score |
|--------|----------|-----------|--------|----------|
| CNN | 0.95 | 0.91 | 0.89 | 0.90 |
| LSTM | 0.96 | 0.92 | 0.88 | 0.89 |
| BERT | 0.95 | 0.91 | 0.87 | 0.88 |
| CNN + LSTM | 0.97 | 0.93 | 0.90 | 0.91 |
| CNN + LSTM + ATTENTION | 0.98 | 0.94 | 0.92 | 0.93 |

## 6 DISCUSSION

The results of the experiments clearly show that the proposed hybrid CNN-LSTM-Attention model is highly effective for text-based web content classification. The model demonstrates significant improvements in accuracy and overall classification performance compared to traditional methods and other deep learning-based approaches. The integration of convolutional layers for capturing local text patterns, LSTM for modeling sequential dependencies, and the attention mechanism for emphasizing important features results in a robust and accurate model. Additionally, the model's strong generalization across different dataset partitions makes it well-suited for practical applications involving large-scale and diverse web content.

The results also indicate that the use of pre-trained GloVe embeddings significantly improves the performance of the model by providing rich semantic representations of the text. The use of K-Fold cross-validation ensures that the evaluation is not biased towards specific splits, and the consistent performance across folds highlights the model's robustness.

Additionally, the comparative analysis of different text vectorization techniques (TF-IDF, Word2Vec, GloVe, BERT) provides insights into their strengths and weaknesses. While TF-IDF offers interpretability, it fails to capture contextual meanings, leading to lower performance. Word2Vec and GloVe achieve better results by capturing semantic relationships, with GloVe outperforming the others due to its ability to model word co-occurrences effectively. BERT, although highly effective in understanding context, requires significant computational resources and may not always generalize well with limited training data.

The experimental results indicate that GloVe provides the highest accuracy (0.94) and maintains a strong balance between precision and recall, making it the best-performing vectorization method in this study. However, despite its effectiveness, further research could explore fine-tuned transformer-based models to enhance text-based web content classification.

Furthermore, the importance of text preprocessing techniques such as tokenization, lemmatization, stop-word removal, and bigram formation has been demonstrated. These methods significantly contribute to improving model performance by refining the textual input and enhancing its quality before vectorization.

The integration of the attention mechanism in the CNN-LSTM model further improves classification performance by identifying key words that contribute to the semantic understanding of web content. The addition of this mechanism allows the model to focus on the most relevant parts of the input text, reducing noise and enhancing decision-making accuracy.

Finally, while the proposed model achieves high accuracy and robustness, it is important to consider computational efficiency. The hybrid approach offers a balanced trade-off between accuracy and processing speed, making it suitable for real-time or near-real-time content classification systems. Future work may focus on optimizing the model for lower computational costs while maintaining high classification capabilities, as well as testing its effectiveness on larger and more diverse datasets.

In conclusion, the hybrid CNN-LSTM-Attention model presents a promising approach to tackling the challenges of text-based web content classification, achieving high accuracy and robustness. Further advancements could explore additional feature engineering techniques, alternative embedding methods, or ensemble learning strategies to enhance overall classification performance.

## CONCLUSIONS

A hybrid approach combining CNN, LSTM, and Attention is proposed to improve the accuracy of text-based web content classification. The optimized model demonstrates consistently high performance, achieving strong results on both training and validation datasets. The Attention mechanism plays a crucial role in enhancing feature identification, which refines the model's classification capabilities. The balanced correlation of key evaluation metrics such as Accuracy, Precision, Recall, F1-score, and AUC indicates the model's stability and helps minimize class bias. Additionally, the model excels at generalization by effectively handling both local and global text dependencies, which helps reduce the gap between training and validation accuracy. While the model requires more computational resources than LSTM, it is still significantly faster than BERT, maintaining high recognition quality. Overall, the proposed hybrid model improves deceptive content detection accuracy by approxi-mately 0.19%–0.37% compared to other contemporary approaches, such as those described in [5], [6], [9], [11], and [12], further demonstrating its effectiveness in real-world scenarios.

**The scientific novelty** of the proposed approach lies in the integration of a hybrid CNN-LSTM-Attention architecture for text-based web content classification. This model combines the strengths of Convolutional Neural Networks (CNN) for capturing local linguistic patterns, Long Short-Term Memory (LSTM) for modeling long-range sequential dependencies, and the Attention mechanism for identifying and emphasizing the most informative features of the text. This combination enhances the model's ability to accurately classify diverse and complex web-based textual content. The incorporation of pre-trained GloVe word embeddings further improves semantic understanding by providing dense vector representations that capture word meaning. Additionally, the use of K-Fold cross-validation ensures robust evaluation, reducing the risk of overfitting and improving generalization. This novel hybrid architecture represents a significant advancement in the field of web content classification.

**The practical significance** of the proposed model is evident in its ability to address a critical issue in web content classification-accurately identifying and categorizing various types of web content. With its high accuracy and efficiency, the hybrid CNN-LSTM-Attention model can be directly applied to real-world text-based classification tasks, including sorting relevant and irrelevant web pages, classifying content into predefined categories, and improving content moderation. The model's robust generalization ensures that it can effectively adapt to new content types and domain-specific text without being overfitted to specific data partitions. Additionally, the integration of the Attention mechanism allows the model to focus on the most critical parts of the text, improving interpretability and making the classification process more transparent. These features make the model suitable for use in real-time content classification systems.

**Prospects for further research** are focused on optimizing the model for better computational efficiency, especially in resource-constrained environments, expanding its capabilities to handle multimodal data inputs like images and videos, adapting the model for cross-language and cross-cultural content classification, and investigating its resilience against adversarial attacks while integrating it with other content-based systems to enhance its robustness and effectiveness.

УДК 004.056.5

# ДОСЛІДЖЕННЯ ГІБРИДНОЇ МОДЕЛІ LSTM-CNN-ATTENTION ДЛЯ КЛАСИФІКАЦІЇ ВЕБ-КОНТЕНТУ НА ОСНОВІ ТЕКСТУ

**Кузь М. В.** – д-р техн. наук, професор кафедри інформаційних технологій, Прикарпатський національний університет імені В. Стефаника, Україна.

**Лазарович І. М.** – канд. техн. наук, доцент кафедри інформаційних технологій, Прикарпатський національний університет імені В. Стефаника, Україна.

**Козленко М. І.** – канд. техн. наук, доцент кафедри інформаційних технологій, Прикарпатський національний університет імені В. Стефаника, Україна.

**Пікуляк М. В.** – канд. техн. наук, доцент кафедри інформаційних технологій, Прикарпатський національний університет імені В. Стефаника, Україна.

**Кваснюк А. Д.** – магістр, Прикарпатський національний університет імені В. Стефаника, Україна.

## АНОТАЦІЯ

**Актуальність.** Класифікація веб-контенту на основі тексту відіграє ключову роль у різних завданнях обробки природної мови (NLP), включаючи виявлення фейкових новин, фільтрацію спаму, категоризацію контенту та автоматизовану модерацію. Оскільки обсяг і складність текстових даних в Інтернеті продовжують зростати, традиційні підходи до класифікації – особливо ті, що спираються на ручне створення ознак або поверхневі методи навчання – мають труднощі в уловлюванні тонких семантичних зв'язків і структурної мінливості сучасного веб-контенту. Ці обмеження призводять до зниження адаптивності та поганої здатності до узагальнення на реальних даних. Тому існує чітка потреба в удосконалених моделях, які можуть одночасно навчатися локальним мовним патернам і розуміти ширший контекстуальний зміст веб-тексту.

**Мета роботи** – підвищення точності та узагальнювальних здатностей моделей класифікації веб-контенту на основі тексту шляхом використання передових технік глибинного навчання. Завданням є покращення витягування локальних та глобальних ознак тексту та навчання послідовних залежностей, що дозволить створити більш ефективну та точну модель для класифікації веб-сторінок з урахуванням їх змісту та контексту.

**Метод.** Це дослідження представляє гібридну архітектуру глибокого навчання, яка інтегрує мережі Long Short-Term Memory (LSTM), згорткові нейронні мережі (CNN) та механізм уваги для покращення класифікації веб-контенту на основі тексту. Для подання слів використовуються попередньо навчений вектор GloVe, який зберігає семантичну подібність. Згорткова мережа (CNN) видобуває локальні патерни *n*-грам і лексичні ознаки, в той час як LSTM моделює довготривалі залежності та послідовну структуру. Інтегрований механізм уваги дозволяє моделі вибірково фокусуватися на найважливіших частинах вхідної послідовності. Модель була оцінена за допомогою датасету, що складається з понад 10 000 веб-сторінок на основі HTML, позначених як легітимні або фейкові. Для оцінки стійкості та узагальненості запропонованого рішення використовувалася 5-кратна крос-валідація.

**Результати.** Експериментальні результати показують, що гібридна модель LSTM-CNN-Attention досягла відмінних результатів, з точністю 0,98, точністю (precision) 0,94, відзивом (recall) 0,92 і F1-мірою 0,93. Ці результати перевершують ефективність базових моделей, що спираються лише на CNN, LSTM або трансформерні класифікатори, такі як BERT. Поєднання компонентів нейронних мереж дозволило моделі ефективно захоплювати як дрібні текстові структури, так і ширший семантичний контекст. Крім того, використання векторів GloVe надало ефективне та дієве подання текстових даних, роблячи модель придатною для інтеграції в системи з вимогами до реального часу або майже реального часу.

**Висновки.** Запропонована гібридна архітектура демонструє високу ефективність у класифікації веб-контенту на основі тексту, особливо в завданнях, що вимагають одночасного видобутку синтаксичних ознак та семантичної інтерпретації. Поєднуючи згорткові, рекурентні та засновані на увазі механізми, модель долає обмеження окремих архітектур і досягає покращеного узагальнення. Ці висновки підтримують більш широке використання гібридних підходів глибокого навчання в додатках NLP, особливо там, де потрібно обробляти та класифікувати складні, неструктуровані текстові дані з високою надійністю.

**КЛЮЧОВІ СЛОВА:** класифікація веб-контенту, LSTM-CNN-Attention, глибоке навчання, обробка природної мови, вектори GloVe, класифікація тексту, гібридна модель, моделювання послідовностей.